\documentclass[a4paper,10pt]{article}
\usepackage[round, numbers]{natbib}
\usepackage{amssymb,amsmath,array}
\usepackage{hyperref}
\usepackage{bm}
\usepackage{subcaption}
\usepackage{multirow}
\usepackage{graphicx}
\usepackage{listings}
\usepackage[a4paper, left=2cm, right=2cm, top=3cm, bottom=2cm]{geometry}
\usepackage[utf8]{inputenc}
\usepackage[T1]{fontenc}
\usepackage[english]{babel}
\usepackage{csquotes}

\usepackage{footnote}
\setcitestyle{square}
\usepackage{mathtools}
\DeclarePairedDelimiter\abs{\lvert}{\rvert}%

\title{Faces: AI Blitz XIII Solutions}
\author{\small\textbf{Andrew Melnik} \and\hspace{-1em}\small\textbf{Eren Akbulut} \and\hspace{-1em}\small\textbf{Jannik Sheikh} \and\hspace{-1em}\small\textbf{Kira Loos} \and\hspace{-1em}\small\textbf{Michael Büttner} \and\hspace{-1em}\small\textbf{Tobias Lenze}}
\date{{\small authors are listed alphabetically}\\\vspace{5mm}{\large Bielefeld University, Germany}}

\begin{document}

\pagestyle{myheadings}
%AIcrowd AI BLITZ XIII
    
\maketitle

\begin{abstract}
\textit{AI Blitz XIII Faces} challenge hosted on \href{http://www.aicrowd.com}{www.aicrowd.com} platform consisted of five problems:\\Sentiment Classification, Age Prediction, Mask Prediction, Face Recognition, and Face De-Blurring.\\Our team \textit{GLaDOS} took second place. Here we present our solutions and results. Code implementation: \href{https://github.com/ndrwmlnk/ai-blitz-xiii}{https://github.com/ndrwmlnk/ai-blitz-xiii}
\end{abstract}
% The code implementation is available here:

\section{Sentiment Classification: Classify Facial Expressions} % Unnumbered section

\subsection{Problem Statement and Dataset} % Numbered section

The goal of the Sentiment Classification problem \cite{AIBlitzXIII-SentimentClassification} of the \textit{AI Blitz XIII Faces} challenge \cite{AIcrowd2022Blitz} was to train a model that can classify faces into three sentiment categories: \textit{negative}, \textit{neutral} and \textit{positive}. The dataset doesn't contain real images but 512-dimensional embedding vectors generated from images of faces by model such as ResNet18 or VGG16. The embedding vectors were extracted from a specific layer of such a model. The original images or the exact model from which the embedding vectors were derived were not specified.

Each face embedding consists of a 512-dimensional vector (see Table \ref{fig:dataset}). The values of the embedding vectors are between 0 and 11.24, while the mean value of the embedding vectors is 0.73. The dataset is divided into three parts: a training set consisting of 5000 samples, a validation set consisting of 2000 samples, and a test set consisting of 3001 samples. The predictions on the test set were evaluated. The metrics used for evaluation are F1 score \cite{sklearnf1} (average=weighted) as a primary score and accuracy score \cite{sklearnacc} as a secondary score.

\begin{table}[ht]
\centering
  \caption{Example of 5 samples from the training dataset. Each embedding has 512 dimensions.}
  \label{fig:dataset}
\begin{tabular}{c}
  \begin{lstlisting}[language=python]
  
id   embedding (512-dim)     label
0  [0.32, 0.98, 1.04, ...] [positive]
1  [0.05, 1.07, 0.60, ...] [negative]
2  [0.41, 0.45, 1.39, ...] [negative]
3  [0.43, 0.19, 0.83, ...] [positive]
4  [0.63, 0.83, 0.39, ...] [neutral]
\end{lstlisting}
\end{tabular}
\end{table}

\subsection{Methods}

See more details on our implementation in the Sentiment Classification notebook \cite{AIBlitzXIIISolutions}. We tested the following approaches: Random Forest Classifier (RFC), K-Nearest Neighbours, Gaussian Naive Bayes, LightGBM, Neural Networks and Support Vector Machines (SVM). For LightGBM and SVM we used Randomized Grid Search with cross-validation. An error-based method for automatic assignment of data subsets to the ensemble NNs using the loss profile of the ensemble is another method for potential improvement of the prediction results \cite{bach2020error}. Out of all the models, Support Vector Classifier (SVC) gave the best results. It has the following parameters: polynomial kernel with a degree of 3, a C-value of 1, and a scaled gamma with a value of $1 / (n_{features} * X.var())$. The multi-class classification is performed using a one-vs-rest approach. Normalization with l1 norm for predictions was the preprocessing method that improved results, albeit by a small amount (see Table \ref{tbl_results}).

\begin{table}[ht]
\centering
\caption{F1 score and accuracy of the baseline Random Forest Classifier and the Support Vector Classifier without preprocessing and with normalization with l1 norm. Trained on the training set and tested on the validation set.}
\begin{subtable}[b]{.6\linewidth}
\centering
\begin{tabular}{ |c||c|c||c|c|   }
 \hline
 \multirow{2}{*}{Dataset} &
      \multicolumn{2}{c||}{F1-Score} &
      \multicolumn{2}{c|}{Accuracy}  \\
   & RFC  & SVC & RFC  & SVC\\
 \hline
 without preprocessing   & 0.668  &  0.808 & 0.673  &  0.800\\
 with l1-norm normalization&  0.679  & 0.807 &0.687  & 0.808\\
 \hline

\end{tabular}
\end{subtable}

\label{tbl_results}
\end{table}

\subsection{Results}

For the predictions, we used the SVC with l1 normalization. We used tenfold cross-validation by combining the training set and the validation set, shuffling them and normalizing them. In this way, we obtained ten estimators. Each of the estimators can make predictions on the test set. We calculated the mean for each sample and rounded it to the nearest label. Omitting certain estimators has an effect on the predictions and consequently on the score. We therefore tried different combinations of the estimators until we found a combination that gave the best results. The final F1 and Accuracy Score on the test set was 0.806 and 0.806 respectivly.

\section{Age Prediction: Predict Age From A Picture}

\subsection{Problem Statement and Dataset}

\begin{figure*}[ht!]
            \includegraphics[width=.2\linewidth]{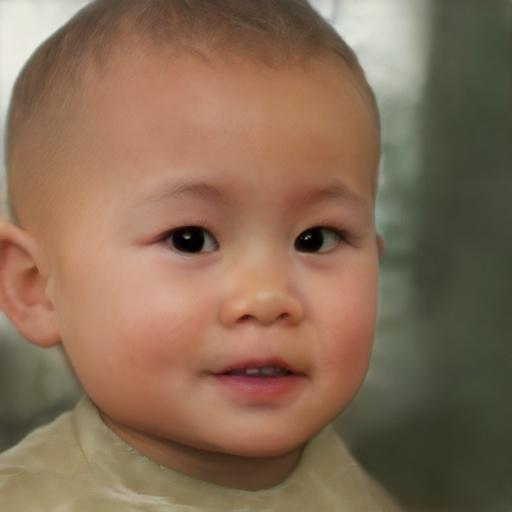}\hfill
%            \caption{Image labeld 0-10}
            \includegraphics[width=.2\textwidth]{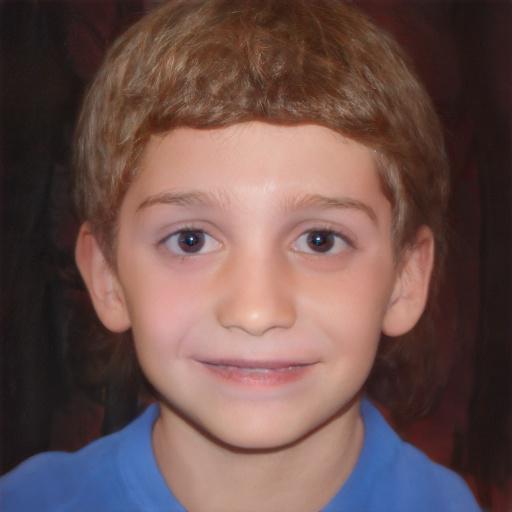}\hfill
%            \caption{Image labeld 10-20}
            \includegraphics[width=.2\textwidth]{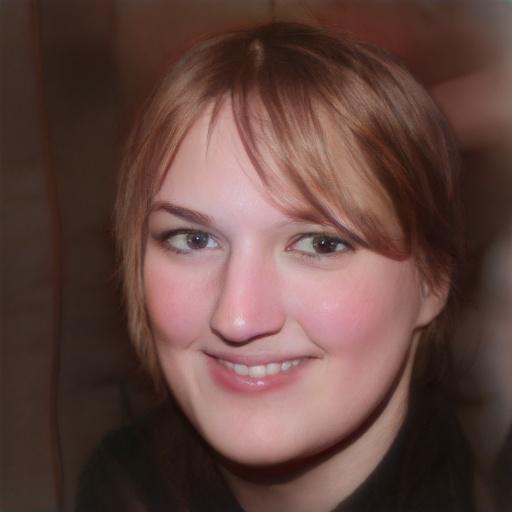}\hfill
%           \caption{Image labeld 20-30}
            \includegraphics[width=.2\textwidth]{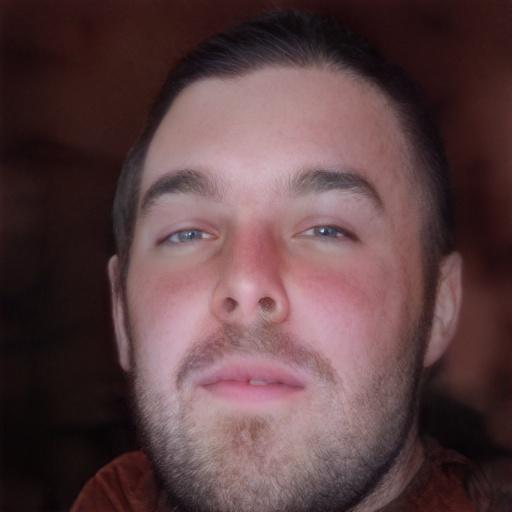}\hfill
%            \caption{Image labeld 30-40}
            \includegraphics[width=.2\textwidth]{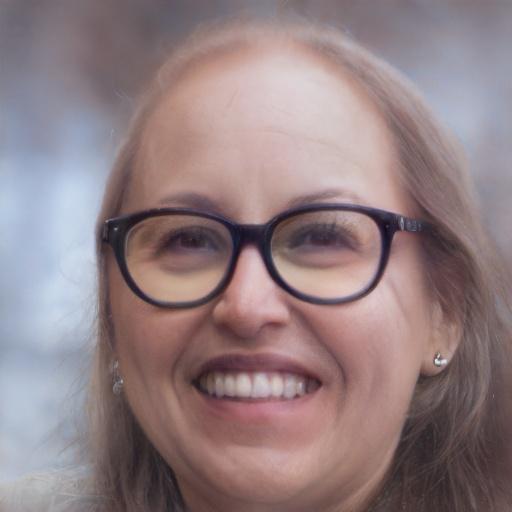}\\
%            \caption{Image labeld 40-50}
            \includegraphics[width=.2\linewidth]{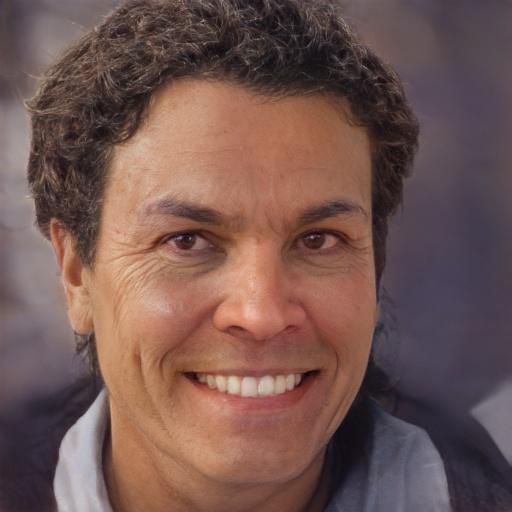}\hfill
%            \caption{Image labeld 50-60}
            \includegraphics[width=.2\textwidth]{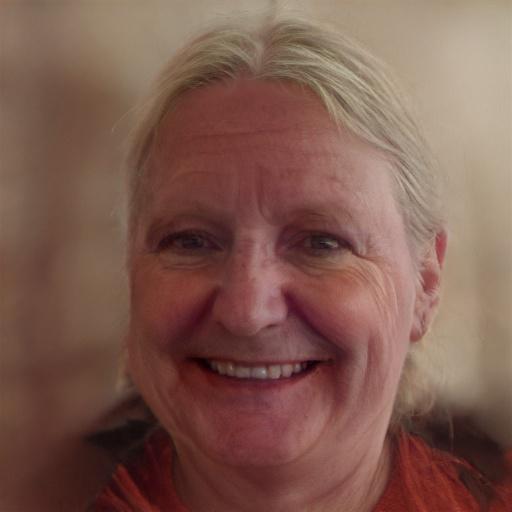}\hfill
%            \caption{Image labeld 60-70}
            \includegraphics[width=.2\textwidth]{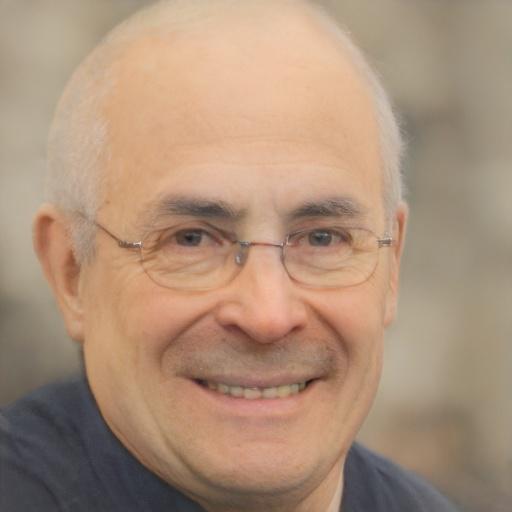}\hfill
%            \caption{Image labeld 70-80}
            \includegraphics[width=.2\textwidth]{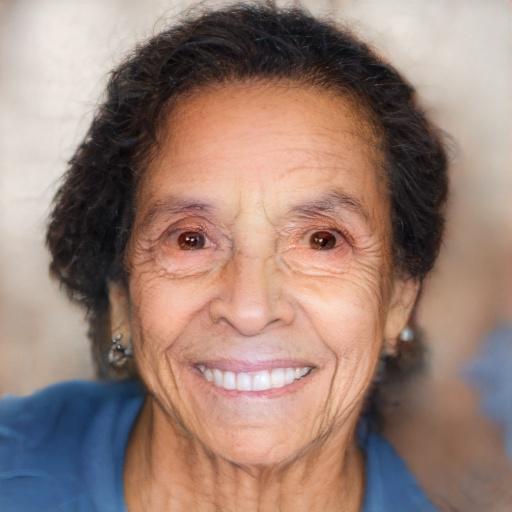}\hfill
%            \caption{Image labeld 80-90}
            \includegraphics[width=.2\textwidth]{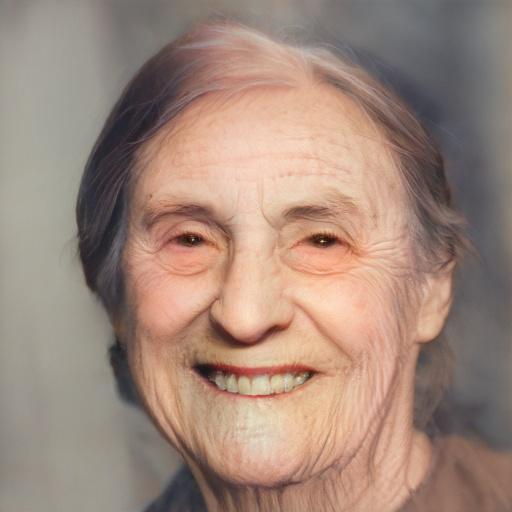}
\caption{Images from train set for each label, ascending. Upper right starting with label 0-10. Lower right starting with label 50-60.}
\label{fig:grid}
\end{figure*}

\begin{figure}[ht]
  \centering     
   \includegraphics[width=0.7\linewidth]{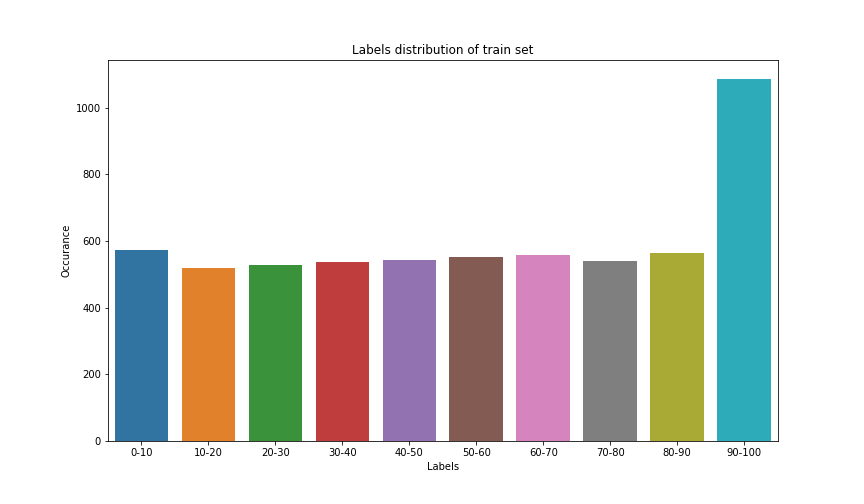}
  \caption{Distribution of train data, after merging}
  \label{fig:distribution}
\end{figure}

The goal of the Age Prediction problem \cite{AIBlitzXIII-AgePrediction} of the \textit{AI Blitz XIII Faces} challenge \cite{AIcrowd2022Blitz} was to predict the age of a human face from an input image. The labels are categorical with 10 classes (0-10 years old, 10-20, 20-30, 30-40, 40-50, 50-60, 60-70, 70-80, 80-90, 90-100). See Figure \ref{fig:grid} for examples. The data is divided into training, validation, and testing sets. The train set contains 4000, the validation set 2000 and the test set 3000 images. The labels were given in training and validation parts of the dataset. The data distribution is shown on Figure \ref{fig:distribution}.

\begin{figure}[!ht]
\centering
\begin{subfigure}{.5\textwidth}
  \centering
  \includegraphics[width=.4\linewidth]{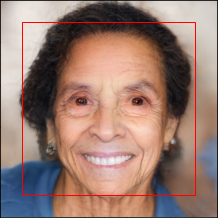}
  \caption{age label 80-90, bounding box}
  \label{fig:bbox1}
\end{subfigure}%
\begin{subfigure}{.5\textwidth}
  \centering
  \includegraphics[width=.4\linewidth]{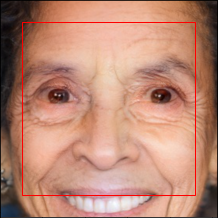}
  \caption{age label 80-90, cropped, bounding box}
  \label{fig:cropped1}
\end{subfigure}
\begin{subfigure}{.5\textwidth}
  \centering
  \includegraphics[width=.4\linewidth]{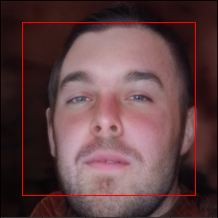}
  \caption{age label 30-40, bounding box}
  \label{fig:bbox2}
\end{subfigure}%
\begin{subfigure}{.5\textwidth}
  \centering
  \includegraphics[width=.4\linewidth]{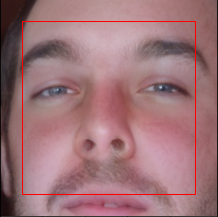}
  \caption{age label 30-40, cropped, bounding box}
  \label{fig:cropped2}
\end{subfigure}
\begin{subfigure}{.5\textwidth}
  \centering
  \includegraphics[width=.4\linewidth]{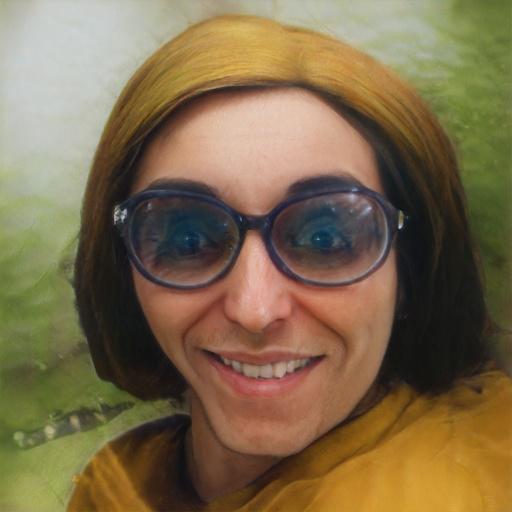}
  \caption{age label 40-50, cartoonish quality}
  \label{fig:noise1}
\end{subfigure}%
\begin{subfigure}{.5\textwidth}
  \centering
  \includegraphics[width=.4\linewidth]{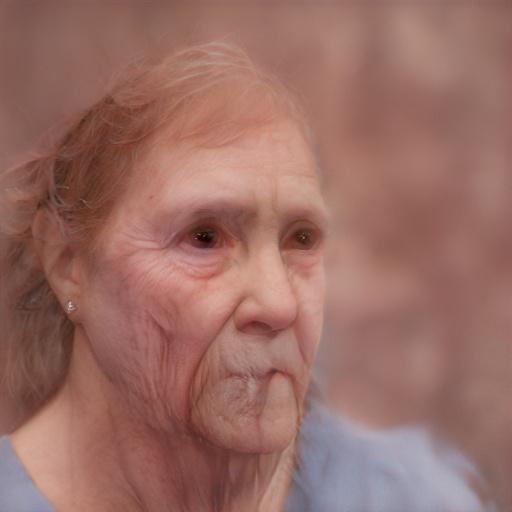}
  \caption{age label 90-100, low quality}
  \label{fig:noise2}
\end{subfigure}%
\caption{Example pictures from the dataset with provided age labels.}
\label{fig:examplefigures}
\end{figure}

Images of human faces in the dataset where generated by a StyleGAN3 model \cite{Karras2021}, which was trained using the Flickr-Faces-HQ dataset. All images are colour images (RGB) of size 512x512x3. Since the images were generated artificially, some images have unnatural artifacts of human faces. Figures \ref{fig:noise1} and \ref{fig:noise2} give you an example.

\subsection{Evaluation Criteria}\label{eval ap}
For the evaluation of models on the leader board \cite{AIcrowd2022Blitz}, weighted average F1 score \cite{sklearnf1} was used as the primary score and the accuracy score \cite{sklearnacc} as the secondary score. The best score for the F1 score is 1 and the worst score is 0. $$F1=2*\frac{precision*recall}{precision+recall}.$$ 

\subsection{Initial experiments}\label{experiments ap}
In this section, we present some of our initial experiments before discussing our best model in \ref{yolo}. 
We started with Random Forest Classifier with RandomizedSearchCV for hyperparamter tuning as well as Light Gradient Boosting Classifier without proper hyperparamter tuning for a subset of the train and test data. The results were not that promising. We achieved a F1 score of 0.33 on the test set with the RFC and even less with the LGBMClassifier.

\subsection{Methods}\label{yolo}
See more details on our implementation in the Age Prediction notebook \cite{AIBlitzXIIISolutions}. We have achieved good results in age prediction using the You Only Look Once (YOLO) architecture \cite{limberg2022yolo, yolohow1, redmon2016look}. Although YOLO is an object detection algorithm, we mainly utilized its class prediction capabilities and abandoned the bounding box prediction functionality. The code implementation of YOLOv5 can be downloaded from GitHub \cite{yolohow1_1}. Since we had no information regarding the bounding boxes, we create fixed bounding boxes 410x410 around the center of the images (x, y = 0.5; width, height = 0.8). Figures \ref{fig:bbox1} and \ref{fig:bbox2} give you a visualization. We trained the model to predict the age class. We also merged the training and validation sets to get more images for training. Additionally, for every image we created a cropped fixed size version, doubling our training set.  We took central 280x280 patch of 512x512 images and resized these 280x280 patches back to 512x512. Examples can be found on Figure \ref{fig:cropped1} and \ref{fig:cropped2}. We trained YOLOv5 with network size L for 200 epochs and used a batch size of 16. Network size L has 46.5M parameters. The different network sizes and characteristics can be found on Github \cite{yolohow1_1}. The training time was about 9 minutes per epoch using Google Colab Pro with a TESLA GPU with 16GB RAM. 

\subsection{Results}
Our result for the Age Prediction problem with the mentioned method and hyperparamters was a F1 score of 0.870 and an accuracy score of 0.871.

\section{Mask Prediction: Predict Mask \& Bounding Box From Images}

\subsection{Problem Statement and Dataset}

\begin{figure}[hb]
  \centering     
  \includegraphics[width=0.5\linewidth]{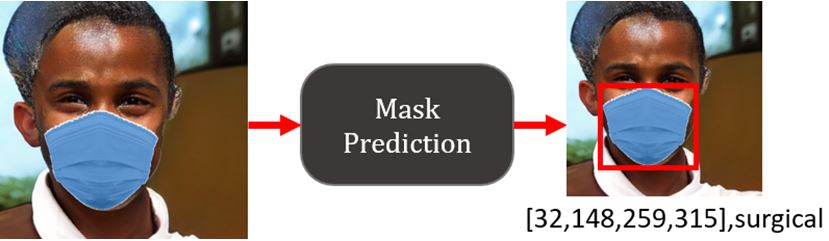}
  \caption{Input: a picture of a person with a mask. Output: the bounding box coordinates and the mask type.}
  \label{fig:MaskPrediction}
\end{figure}

\begin{figure}[hb]
  \centering     
  \includegraphics[width=0.5\linewidth]{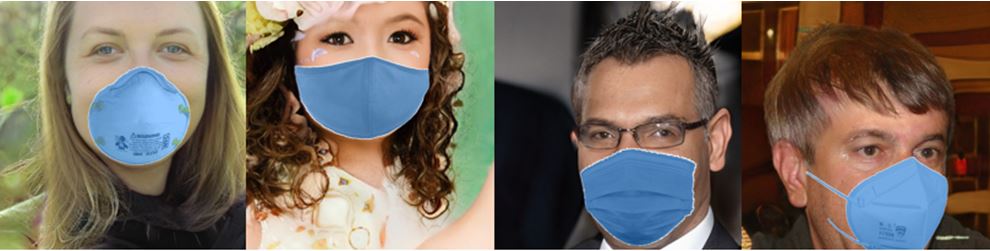}
  \caption{Samples of the dataset for each mask type. L.t.r.: N95, cloth, surgical, KN95.}
  \label{fig:Masktypes}
\end{figure}

The goal of the Mask Prediction problem \cite{AIBlitzXIII-MaskPrediction} of the \textit{AI Blitz XIII Faces} challenge \cite{AIcrowd2022Blitz} was to predict the type and bounding box of the mask that a person is wearing in an image (fig.\ref{fig:MaskPrediction}). Trained on the provided dataset, we achieved an accuracy of 99.3\%, which was the top score in this problem.

\begin{figure}[ht]
  \centering     
  \includegraphics[width=0.5\linewidth]{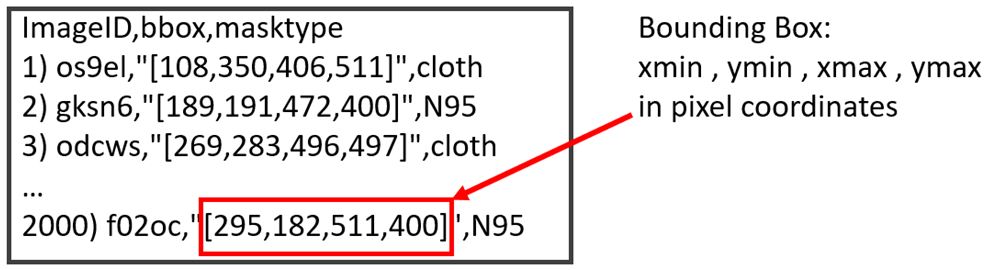}
  \caption{Data structure of CSV files: ImageID, bounding box and mask type. Also the format used for the bounding box on the right.}
  \label{fig:DataStructure}
\end{figure}

The provided dataset was divided into three different sets. Train, validation and test sets. Each set contains 5000, 2000, and 3000 512x512pixel images, respectively. Each image contains one of 4 mask types: Surgical, N95, KN95 and Cloth (fig.\ref{fig:Masktypes}). CSV files were also provided for the training and validation images, which contained information about the mask type for each image and the corresponding bounding box of the mask in pixel coordinates (fig.\ref{fig:DataStructure}).

\subsection{Methods}
See more details on our implementation in the Mask Prediction notebook \cite{AIBlitzXIIISolutions}. You Only Look Once (YOLO) \cite{limberg2022yolo} was chosen as the prediction method. YOLOv5 \cite{yolohow1_1} is the fifth iteration of YOLO, a state-of-the-art, real-time object detection system capable of creating bounding boxes around detected objects in images. YOLO requires a specific format and file hierarchy. For this the following folder structure has to be implemented, containing \textit{.../images/train}, \textit{.../images/val}, \textit{.../images/test} images, and \textit{.../labels/train}, \textit{.../labels/val} labels.

Labels in the YOLO format must be created and moved to the respective \textit{labels} folders. These labels are \textit{.txt} files for each image, with the same name as the corresponding images. For example, the image \textit{.../images/train/abc.jpg} must have a counterpart \textit{.../labels/train/abc.txt}. Each label contains the object class (a numeric value starting with 0), and the bounding box in the following format:

<object-class> <x> <y> <width> <height>

It should also be noted that YOLO uses values that are image size related and not pixel related.  To convert the values from the CSV files to YOLO format, the following formulas were used:

\begin{figure}[ht]
  \centering     
  \includegraphics[width=0.5\linewidth]{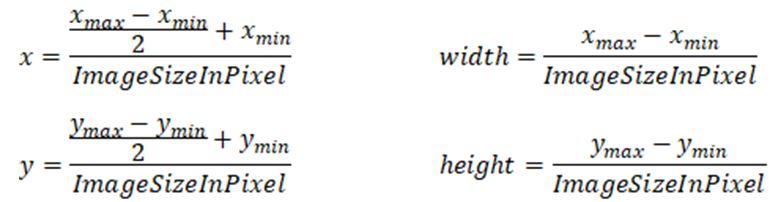}
  \label{fig:Math}
\end{figure}

Training in YOLOv5 can be started by a command with parameters that specify the image size in pixel, the batch size, number of epochs, the size of YOLOv5 network on which to train, and the path to the \textit{mask\_prediction.yml} \cite{AIBlitzXIIISolutions}. The \textit{mask\_prediction.yml} file contains the paths to the training and validation images as well as the number and names of classes \cite{AIBlitzXIIISolutions}. The training time per epoch for the training set of 3000 images and a batch size of 16 is about 15 minutes on a Tesla K80 via Google Colab. 

To get the bounding box coordinates, the --save-txt parameter is required to generate labels in YOLO format. These bounding box coordinates have to be converted into the format of the CSV files to be submitted for scoring. The scoring is based on the Average Precision of mask type and bounding box ( @ IoU=0.50:0.50 )

\subsection{Results}
The prediction with the highest confidence score for each image was selected as the predicted object. The best result, with 99.3\% accuracy, was achieved after training for 100 epochs, with a batch size of 16 images and with the YOLOv5l model.

% \begin{figure}[ht]
%   \centering     
%   \includegraphics[width=0.6\linewidth]{imgs/MP/Table.JPG}
%   \label{fig:Table}
%   \caption{Overview over a few combinations for the number of epochs, batch size and the resulting score.}
% \end{figure}

\section{Face Recognition: Find The Correct Face In Crowd}

\subsection{Problem Statement and Dataset}

\begin{figure}
% figure 1
\centering
\begin{subfigure}{.25\linewidth}
\centering
\includegraphics[width=\textwidth]{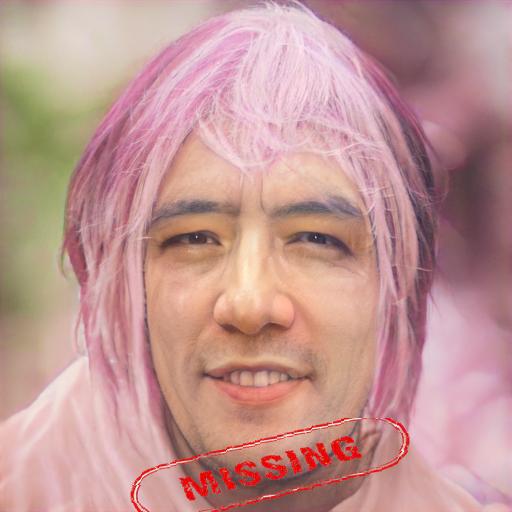}
\end{subfigure}
\begin{subfigure}{.25\linewidth}
\centering
\includegraphics[width=\textwidth]{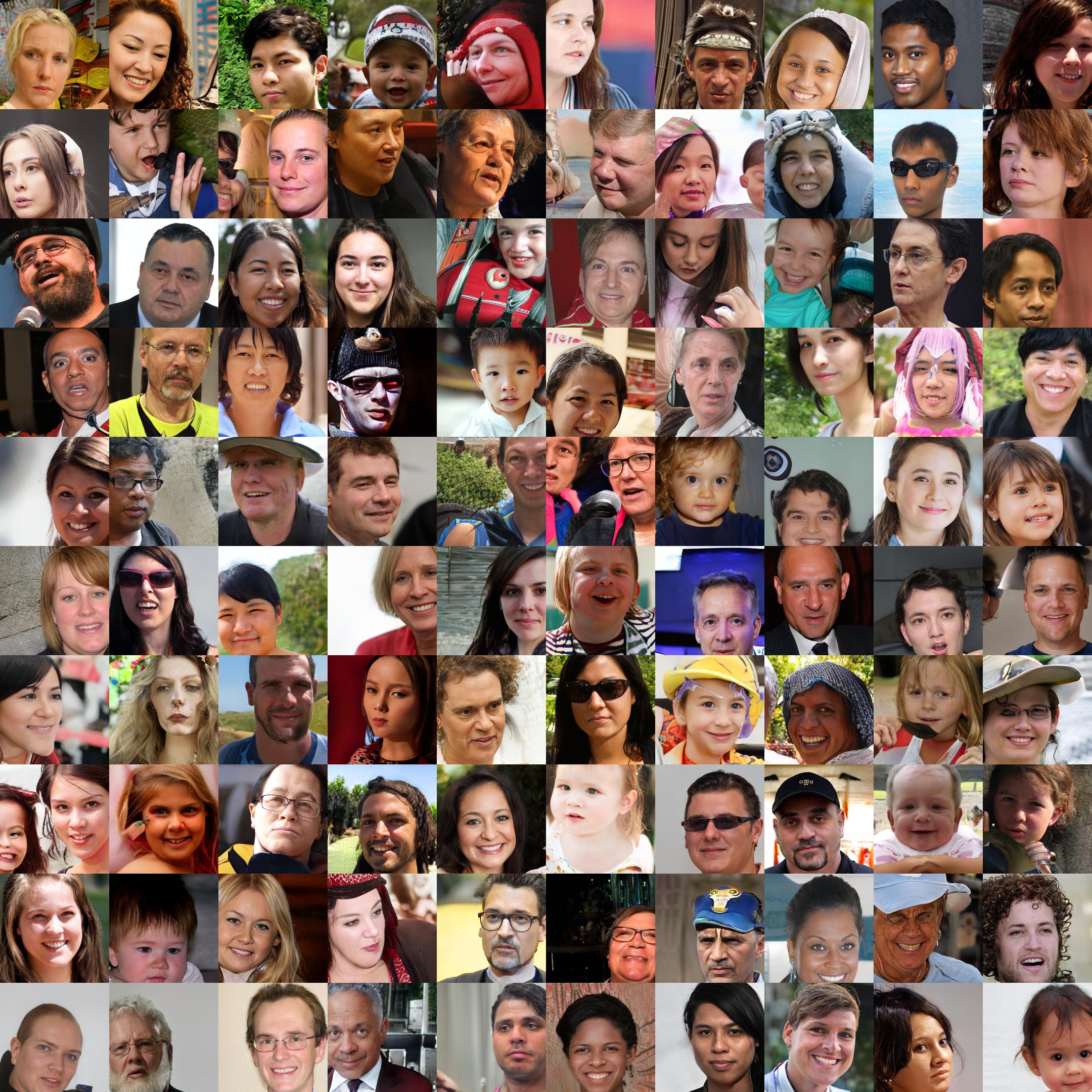}
\end{subfigure}
\begin{subfigure}{.25\linewidth}
\centering
\includegraphics[width=\textwidth]{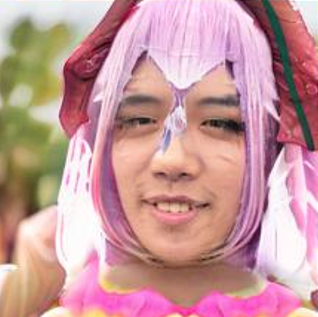}
\end{subfigure}
% figure 2
\centering
\begin{subfigure}{.25\linewidth}
\centering
\includegraphics[width=\textwidth]{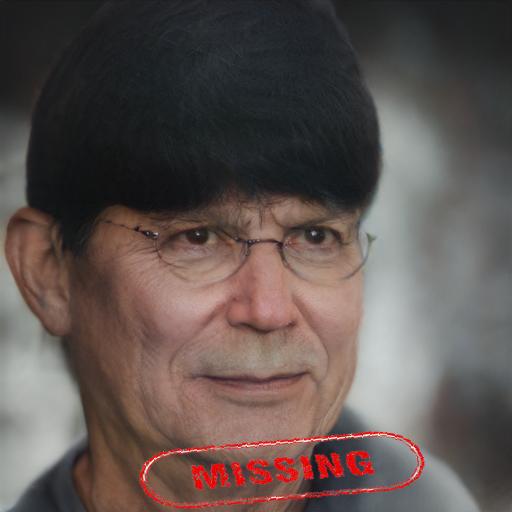}
\end{subfigure}
\begin{subfigure}{.25\linewidth}
\centering
\includegraphics[width=\textwidth]{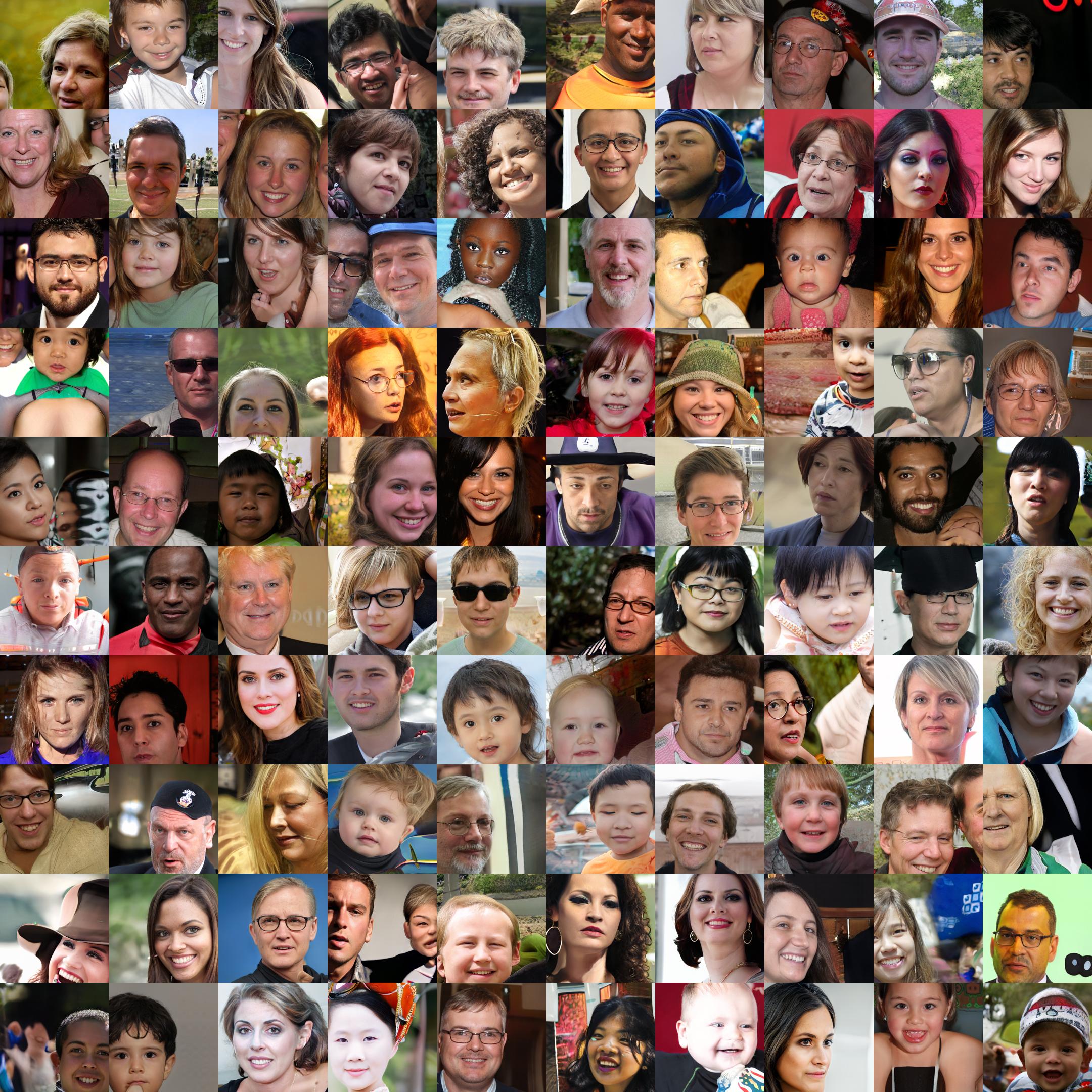}
\end{subfigure}
\begin{subfigure}{.25\linewidth}
\centering
\includegraphics[width=\textwidth]{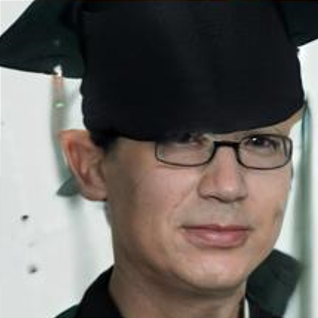}
\end{subfigure}
% figure 3
\centering
\begin{subfigure}{.25\linewidth}
\centering
\includegraphics[width=\textwidth]{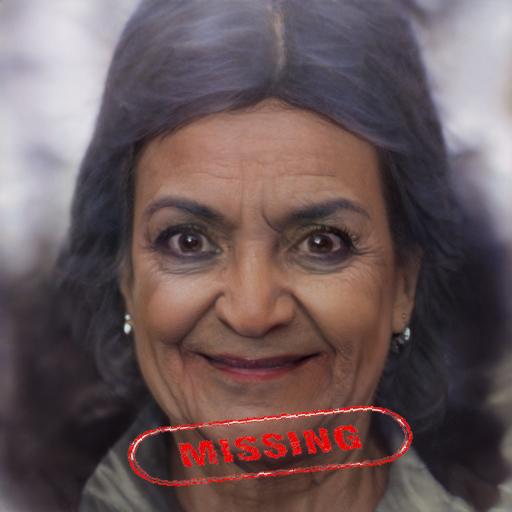}
\end{subfigure}
\begin{subfigure}{.25\linewidth}
\centering
\includegraphics[width=\textwidth]{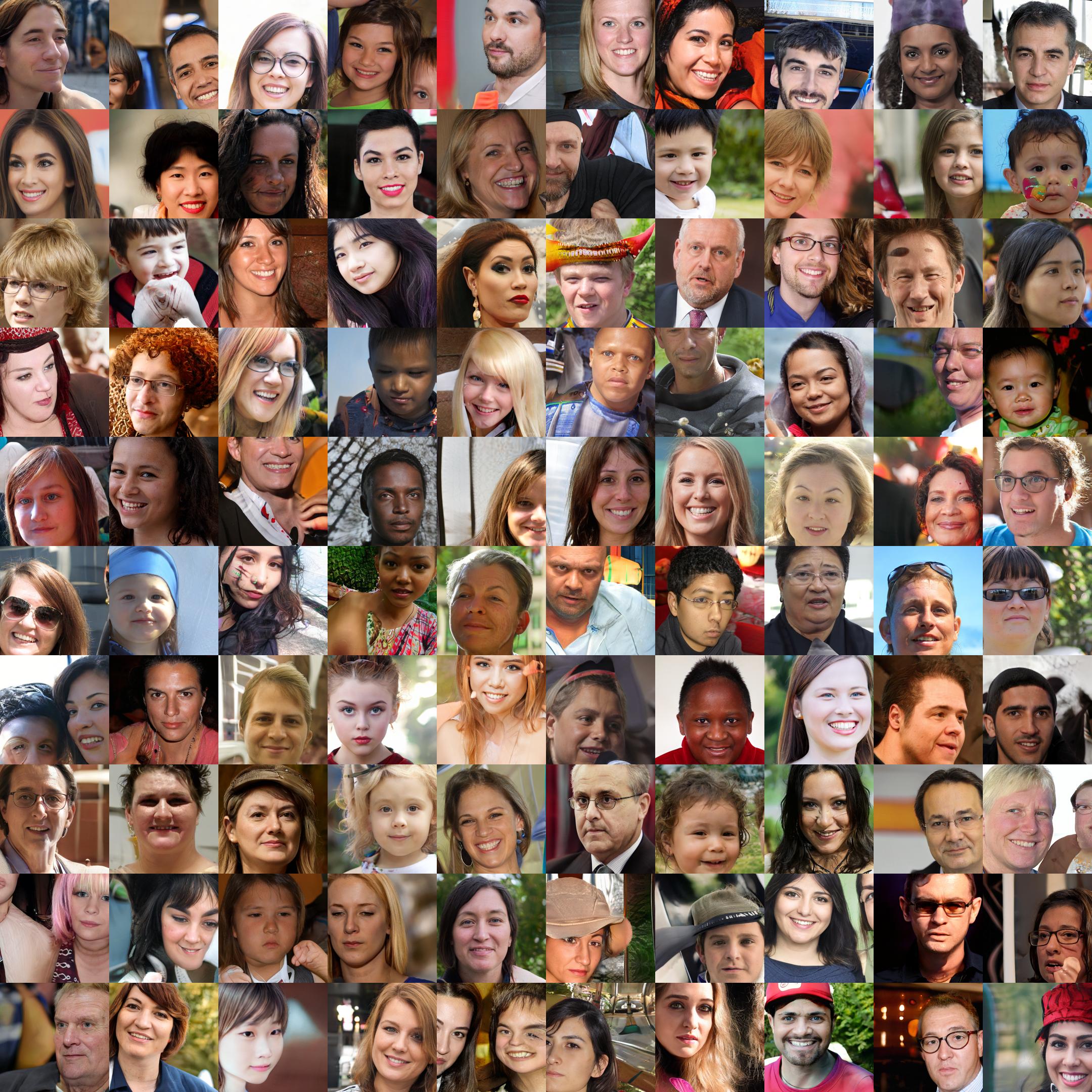}
\end{subfigure}
\begin{subfigure}{.25\linewidth}
\centering
\includegraphics[width=\textwidth]{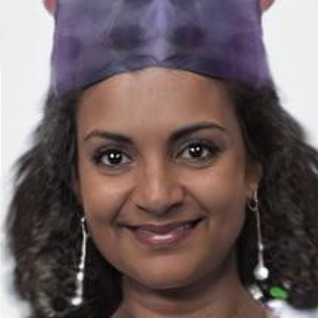}
\end{subfigure}
% figure 4
\centering
\begin{subfigure}{.25\linewidth}
\centering
\includegraphics[width=\textwidth]{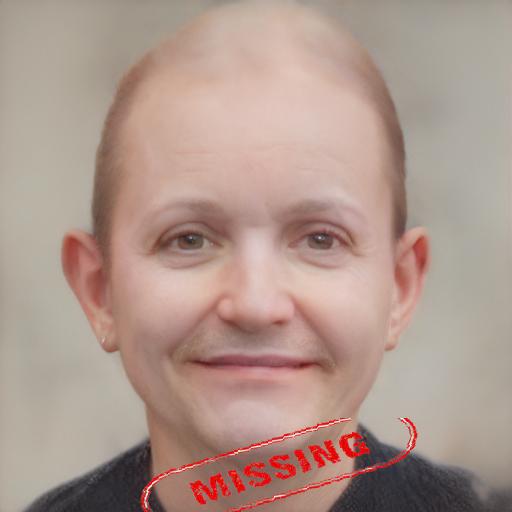}
\end{subfigure}
\begin{subfigure}{.25\linewidth}
\centering
\includegraphics[width=\textwidth]{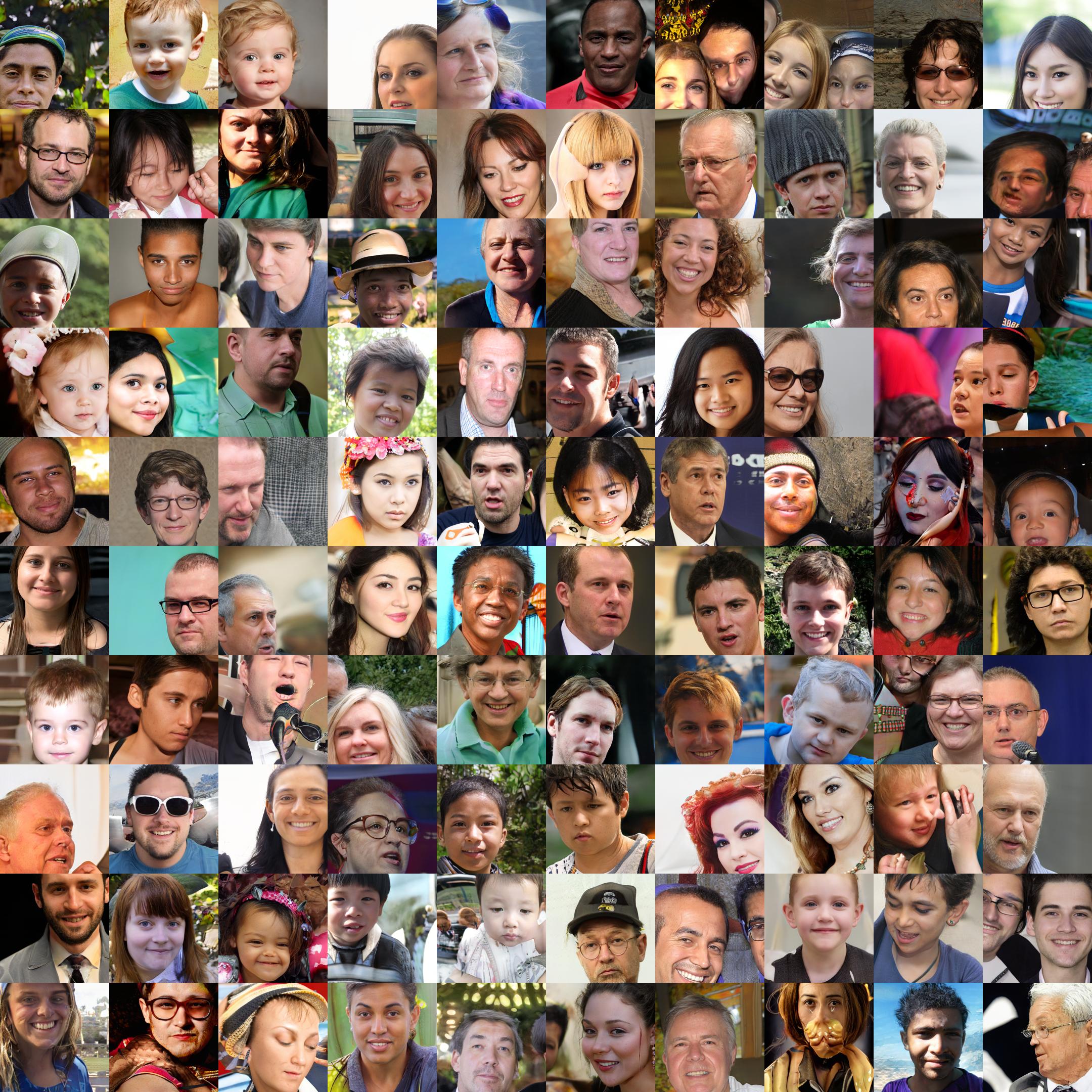}
\end{subfigure}
\begin{subfigure}{.25\linewidth}
\centering
\includegraphics[width=\textwidth]{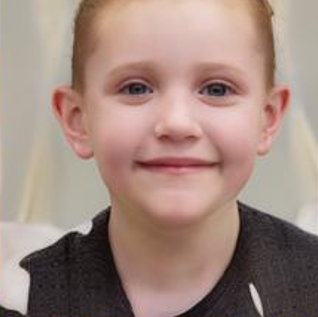}
\end{subfigure}
% figure 5
\centering
\begin{subfigure}{.25\linewidth}
\centering
\includegraphics[width=\textwidth]{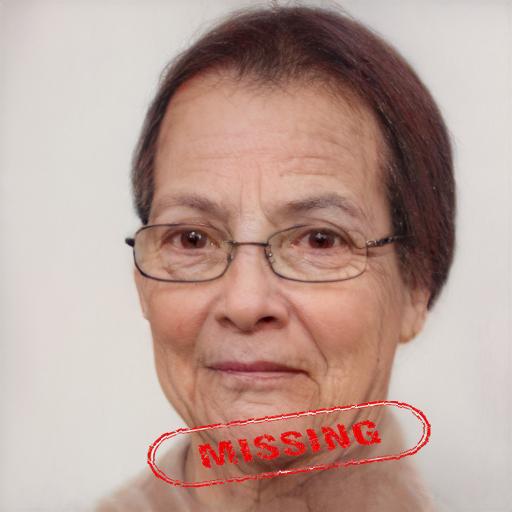}
\end{subfigure}
\begin{subfigure}{.25\linewidth}
\centering
\includegraphics[width=\textwidth]{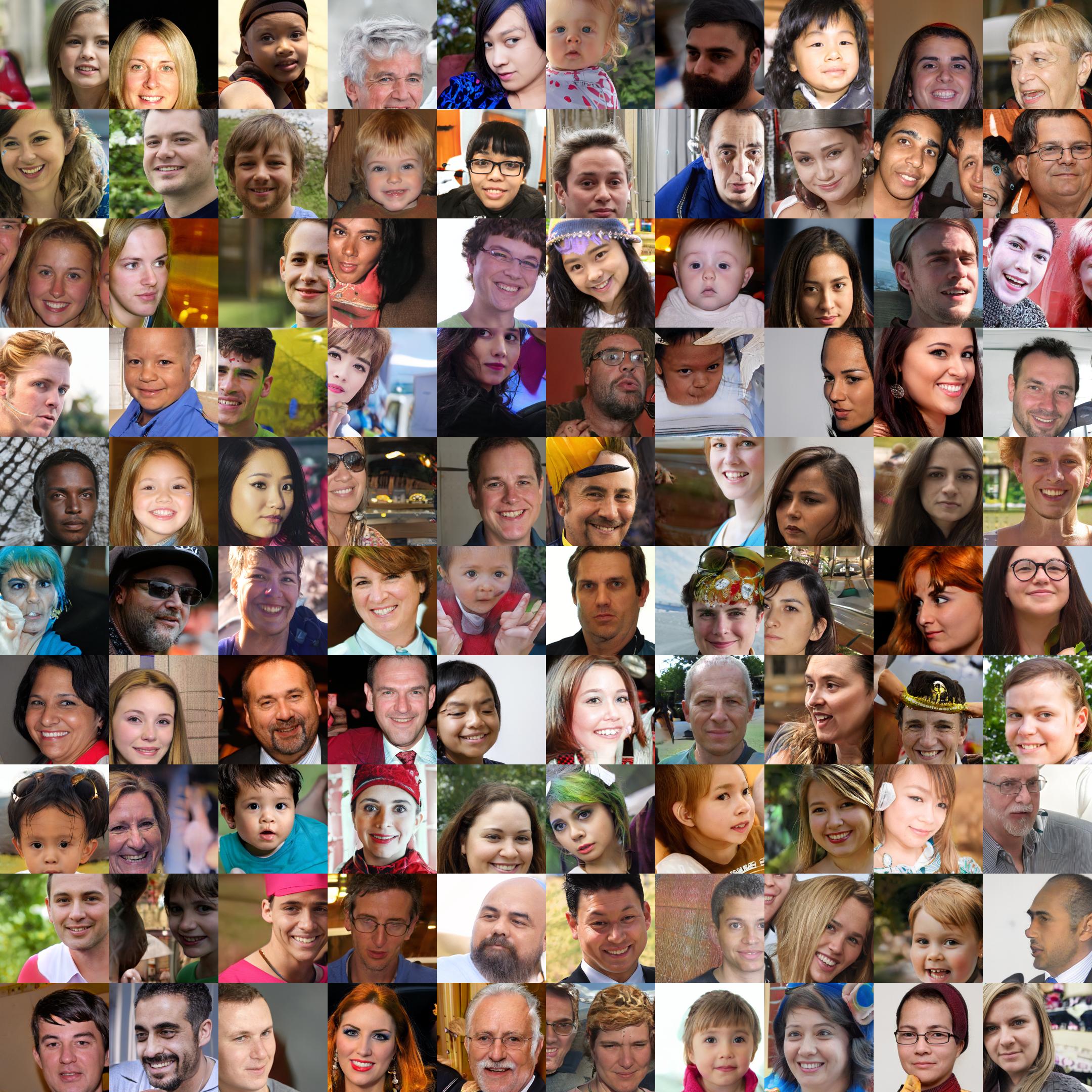}
\end{subfigure}
\begin{subfigure}{.25\linewidth}
\centering
\includegraphics[width=\textwidth]{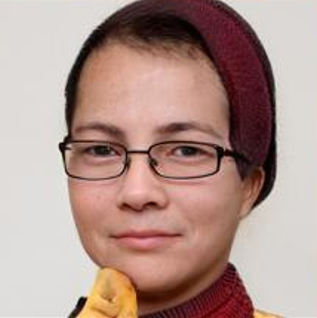}
\end{subfigure}
% end of figure 5
\caption{Missing image (left), target image with 100 faces (middle) and prediction (right) for a set of images from the dataset.}
\label{fig:examples}
\end{figure}

The goal of the Face Recognition problem \cite{AIBlitzXIII-FaceRecognition} of the \textit{AI Blitz XIII Faces} challenge \cite{AIcrowd2022Blitz} was to find the \textit{missing} face in crowd (target image). The missing image contains a single face that has to be found in the corresponding target image. Each target image shows 100 faces. The unlabeled dataset \cite{AIBlitzXIII-FaceRecognition} consists of 1000 missing and target image pairs. The faces were all generated using a StyleGAN3 model. The person in the missing image differ in some attributes in the corresponding target image. See examples in Figure \ref{fig:examples}.

%\par\vspace{0.5cm}

\subsection{Methods}
See more details on our implementation in the Face Recognition notebook \cite{AIBlitzXIIISolutions}. Since the provided dataset is unlabeled, we could not train the face detection or face recognition ourselves. We used pre-trained models and methods for face-recognition \cite{faceRecogGithub}\cite{dlibGithub}.

In our first approach, we used face-alignment \cite{faceAligGithub} to compute 68 3D landmarks. We computed the distances between each possible pair of landmarks for each given face. Afterwards we compared these distances between the missing face and every face seen in the target image, selecting the one with the smallest error as the prediction. The distances are normalized by the jaw width, which we defined as the distance between landmark 1 and 17. This got us an accuracy of 22\%, which could not be improved by using only a subset of distances to compare, nor by weighting some distances higher.

In our second approach, we used Google's MediaPipe \cite{mediapipePaper}\cite{mediapipeGithub} to compute 468 3D landmarks in a given face. Initially, we only used the landmark groups \cite{mediapipeGroupsGithub} which were provided by Google's MediaPipe, but by using our own groups in addition to that, we could increase the accuracy to 50\%. Our own groups provided landmark paths for the nose and the facial structure.

In our third approach we decided to use a face-recognition method \cite{faceRecogGithub}. It allowed us to use dlib \cite{dlibGithub} library that provides an encoding function to find an embedding for a face, and a distance function to calculate the difference between two encodings. We used two different encoding models to find the encodings: \textit{small} uses 5 landmarks and \textit{large} uses 68 landmarks. To predict the target images, we used the mean over the distances calculated by those two models. This increased our initial accuracy for this approach by 2\%, resulting in our final accuracy: 78.8\%.

\subsection{Results}
For the third approach, we got an accuracy of 78.8\%. Selection of a random image would give us an accuracy of 1\%, since there are 100 possible target images. The prediction time for our notebook was 2.5 hours using Google Colab Pro computation resources.

\section{Face De-Blurring: Convert Blurred image Into Clear Image}

\subsection{Problem Statement and Dataset}

\begin{figure}[ht!]
  \centering     
  \includegraphics[width=1\linewidth]{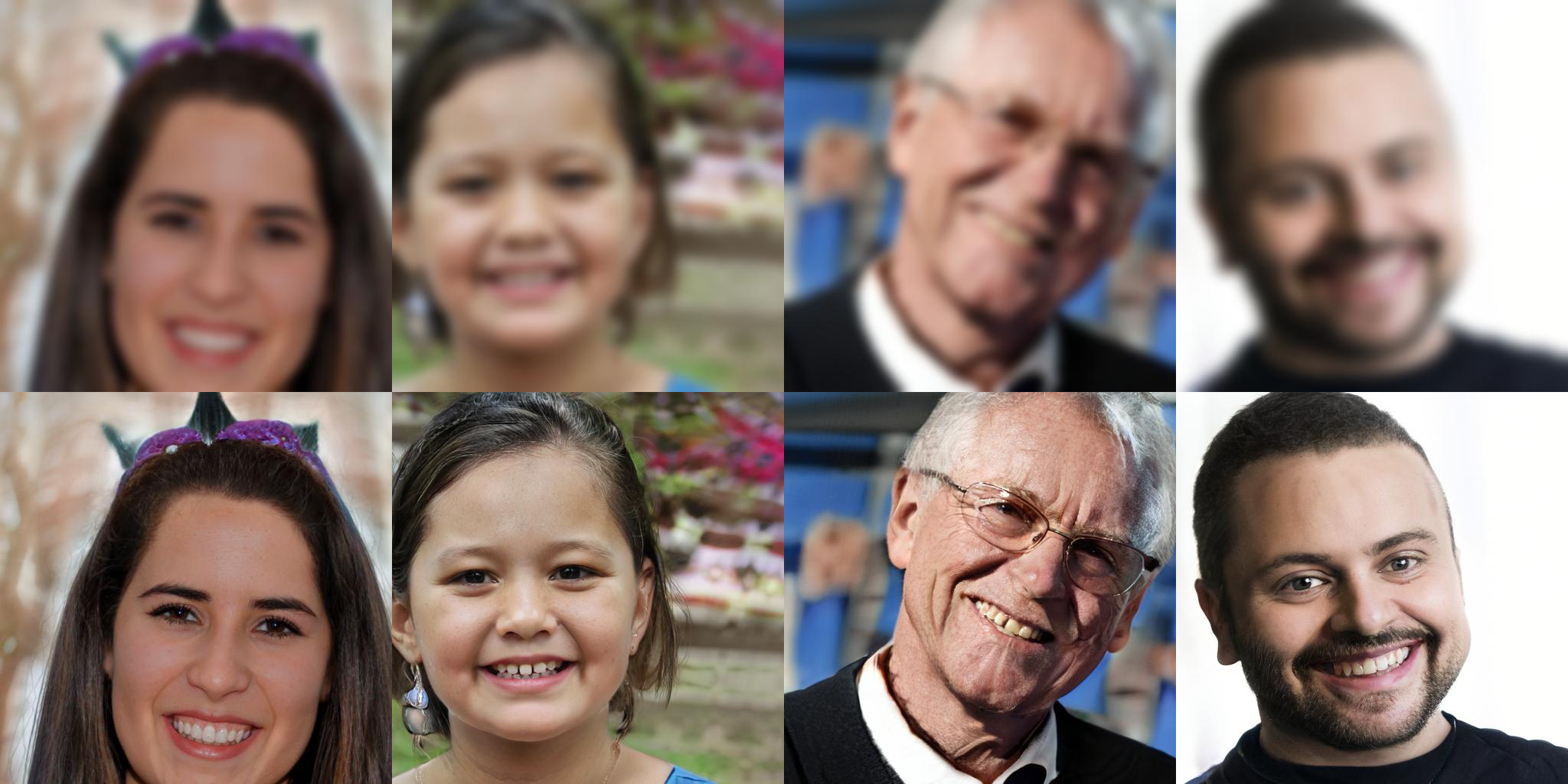}
  \caption{Examples of training pairs: Original (Top) and Blurred (Bottom).}
  \label{fig:extraining}
\end{figure}

The goal of the Face Deblurring problem \cite{AIBlitzXIII-FaceDeBlurring} of the \textit{AI Blitz XIII Faces} challenge \cite{AIcrowd2022Blitz} was to generate a sharp image of a human face from a blurred version of the same human face. The dataset used for this challenge are images of human faces, which were generated by a StyleGAN3 model \cite{Karras2021}, which was trained on the Flickr-Faces-HQ dataset. The data is split into train, validation \& test sets. The train set contains 5000 pairs of blurred and original images, the validation set 2000 pairs of blurred and original images, and the test set 3000 blurred images. All images are colour images (RGB) of size 512x512x3. See example image pairs in Figure \ref{fig:extraining}.

\begin{figure}[ht!]
  \centering     
  \includegraphics[width=1\linewidth]{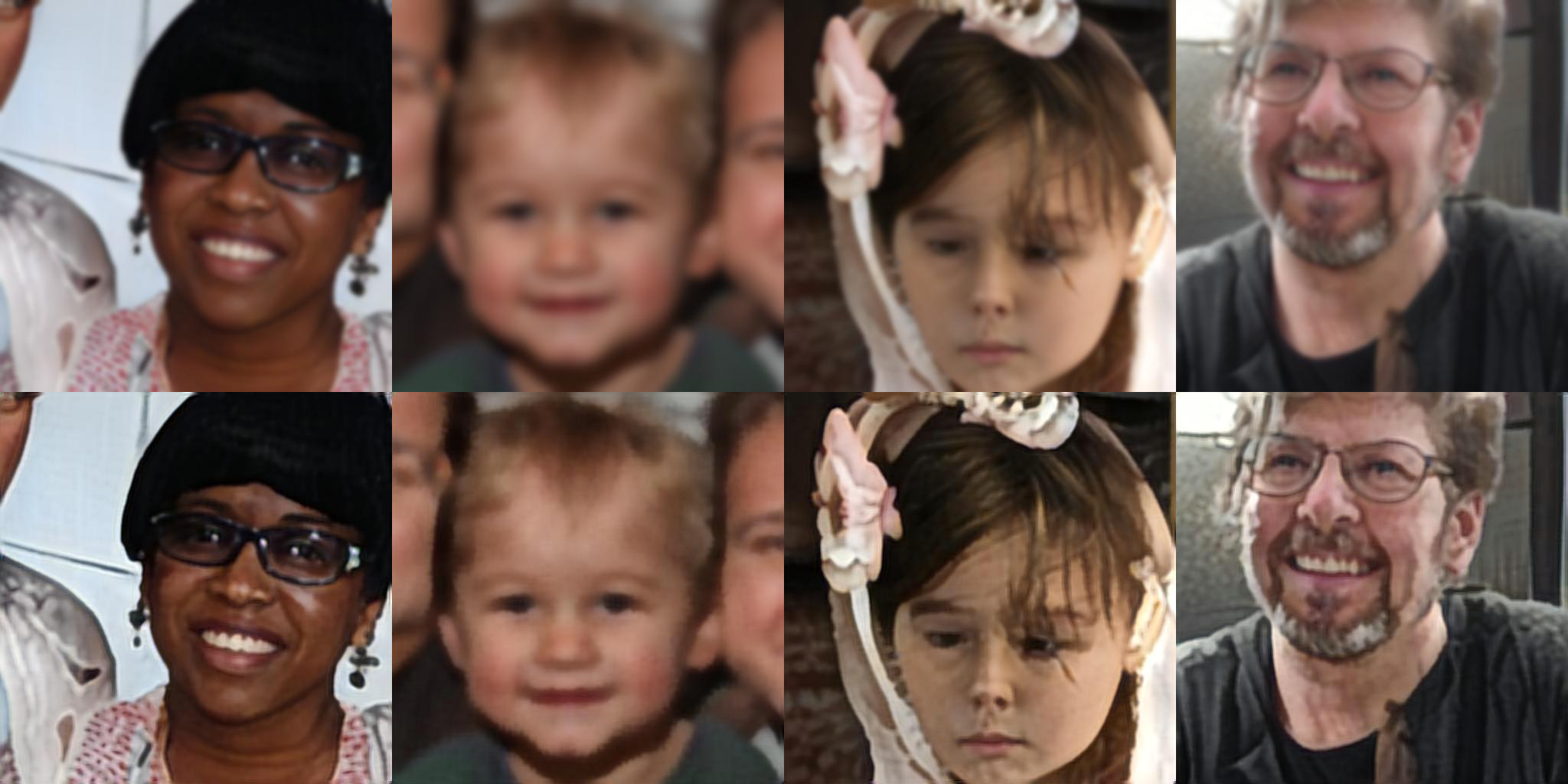}
  \caption{Examples of our results: test blurred image (Bottom) and deblurred image (Top)}
  \label{fig:extest}
\end{figure}

\begin{figure}[ht!]
  \centering     
  \includegraphics[width=0.75\linewidth]{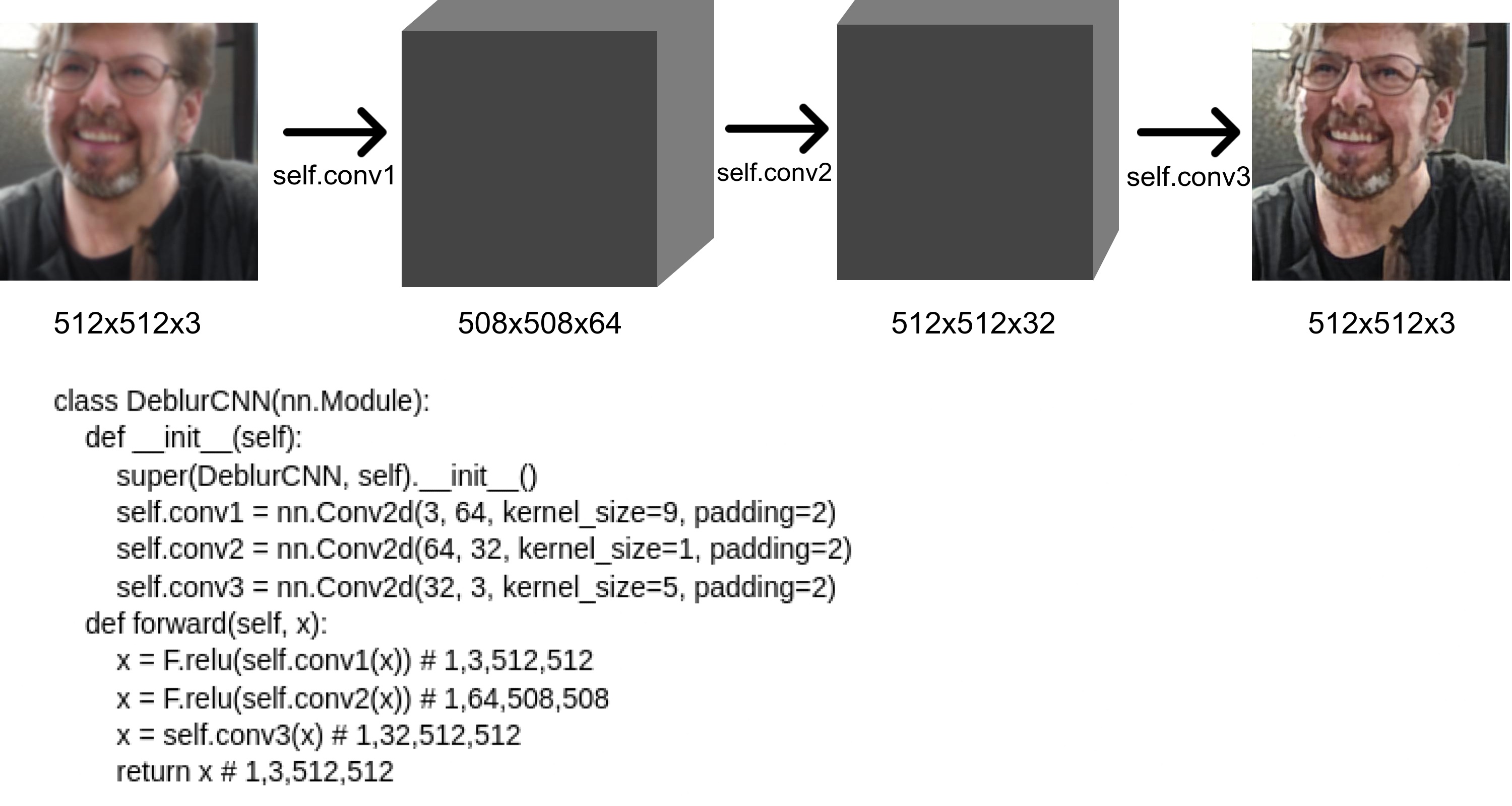}
  \caption{SRCNN Architecture}
  \label{fig:srcnn2}
\end{figure}

\begin{figure}[ht!]
  \centering     
  \includegraphics[width=0.75\linewidth]{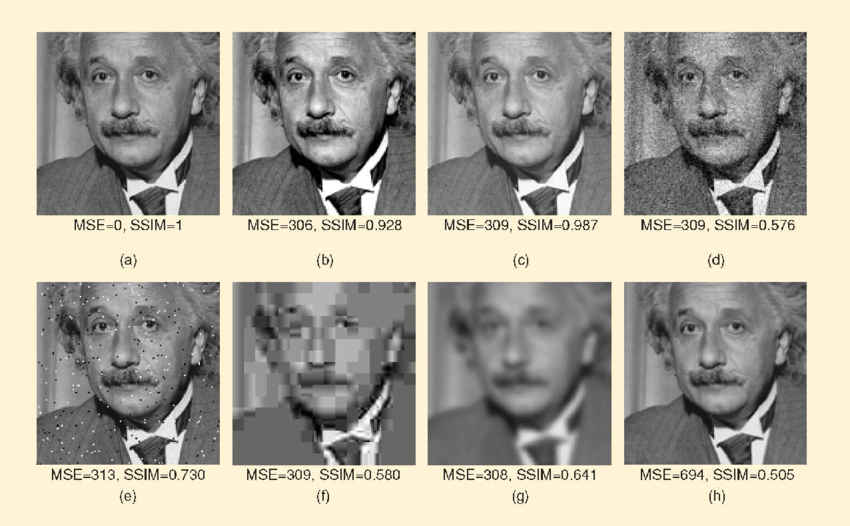}
  \caption{MSE vs. SSIM \cite{phdthesis}}
  \label{fig:compssim}
\end{figure}

% \begin{figure}[ht]
%   \centering     
%   \includegraphics[width=0.9\linewidth]{imgs/FD/srcnn2.jpg}
%   \caption{SRCNN Architecture in Detail \cite{dong2015image}}
%   \label{fig:srcnn2}
% \end{figure}

\subsubsection{Evaluation Criteria}\label{eval fd}
The main evaluation criteria was Structural Similarity (SSIM) and for 2 images "x" and "y" how it's being calculated is shown below. 
\begin{equation}
  \mathit{SSIM}(x,y) = \frac{(2\mu_x\mu_y + C_1) + (2 \sigma _{xy} + C_2)} 
    {(\mu_x^2 + \mu_y^2+C_1) (\sigma_x^2 + \sigma_y^2+C_2)}
  \label{eq:SSMI}
\end{equation}
To solve potential conflicts in case of having an identical SSIM score a second metric was utilized, Peak signal-to-noise ratio (PSNR).
\begin{equation}
\mathit{PSNR}(x,y)=\frac{10\log_{10}[\max(\max(x),\max(y))]^2}{\abs{x-y}^2}
  \label{eq:PSNR}
\end{equation}
For such image reconstruction problems SSIM is able to give more meaningful results than Mean Squared Error (MSE) as you can see in \ref{fig:compssim}.

\subsection{Methods}\label{experiments fd}

\subsubsection{SRCNN}\label{SRCNN}
See more details on our implementation in the Face De-Blurring notebook \cite{AIBlitzXIIISolutions}. We mainly focused on the SRCNN \cite{dong2015image} approach throughout this challenge with a modified loss function. Other approaches \cite{FacesSurvey} \cite{mao2021deep} \cite{wang2021uformer} \cite{zamir2021restormer} required large computing resources, which we did not have. SRCNN is an architecture is end-to-end learning approach that maps low resolution images into high resolution images. This approach is based on a CNN layers and therefore can be adopted to all input sizes of images.

The original SRCNN paper proposes to used a MSE loss only. We combined MSE and L1 losses with a 3rd loss function from SSIM family, Multi-scale structural similarity (MSSSIM) \cite{1292216}. 
The original loss function was  
\begin{equation}
    \mathit{Loss}(x,y)= MSE(x,y)
      \label{eq:loss1}
\end{equation}
and modified into
\begin{equation}
    \mathit{Loss}(x,y)= MSE(x,y) + L1(x,y) + MSSIM(x,y)
      \label{eq:loss2}
\end{equation}
Too avoid MSSIM loss to return NaN from time to time, at each scale of the MSSIM calculation we normalized the output using a Rectified Linear Unit (ReLu) normalizer.

The network only contains 3 CNN layers (see Figure \ref{fig:srcnn2}), the first CNN layer with the kernel size 9 takes a lower resolution image and extracts a set of feature maps, then the second CNN layer with the kernel size 1 maps there features non-linearly to a higher resolution patch representations and the final CNN layer with the kernel size 5 combines the predictions within a spatial neighbourhood to produce the final high-resolution image. See our results in Figure \ref{fig:extest}.

The average training time on a laptop was around 8-12 hours for 120-180 epochs with a batch size of 1, 2 and 4. AdamW with a learning rate of 0.001 used for the final submission. To optimize the training process we used a scheduler with the patience of 4 epochs and factor 0.5.

\subsection{Results}
In this problem we got SSIM of 0.748 and an PSNR of 26.627. See examples of our results in Figure \ref{fig:extest}.

\clearpage
\bibliography{bibliography}

\begin{thebibliography}{10}

\bibitem{AIBlitzXIII-SentimentClassification}
AIcrowd, ``{AI Blitz XIII Faces: Sentiment Classification - Classify Facial
  Expressions}.''
  \url{https://www.aicrowd.com/challenges/ai-blitz-xiii/problems/sentiment-classification};
  accessed April 4, 2022.

\bibitem{AIcrowd2022Blitz}
AIcrowd, ``{AI Blitz XIII Faces}.''
  \url{https://www.aicrowd.com/challenges/ai-blitz-xiii}; accessed April 4,
  2022.

\bibitem{sklearnf1}
``sklearn.metrics.f1score.''
  \url{https://scikit-learn.org/stable/modules/generated/sklearn.metrics.f1_score.html}.
\newblock accessed April 4, 2022.

\bibitem{sklearnacc}
``sklearn.metrics.accuracy\_score.''
  \url{https://scikit-learn.org/stable/modules/generated/sklearn.metrics.accuracy_score.html}.
\newblock accessed April 4, 2022.

\bibitem{AIBlitzXIIISolutions}
A.~Melnik, E.~Akbulut, J.~Sheikh, K.~Loos, M.~Büttner, and T.~Lenze, ``{Faces:
  AI Blitz XIII Solutions}.'' \url{https://github.com/ndrwmlnk/ai-blitz-xiii};
  accessed April 4, 2022.

\bibitem{bach2020error}
N.~Bach, A.~Melnik, F.~Rosetto, and H.~Ritter, ``An error-based addressing
  architecture for dynamic model learning,'' in {\em International Conference
  on Machine Learning, Optimization, and Data Science}, pp.~617--630, Springer,
  2020.

\bibitem{AIBlitzXIII-AgePrediction}
AIcrowd, ``{AI Blitz XIII Face: Age Prediction - Predict Age From A Picture}.''
  \url{https://www.aicrowd.com/challenges/ai-blitz-xiii/problems/age-prediction};
  accessed April 4, 2022.

\bibitem{Karras2021}
T.~Karras, M.~Aittala, S.~Laine, E.~H\"ark\"onen, J.~Hellsten, J.~Lehtinen, and
  T.~Aila, ``Alias-free generative adversarial networks,'' in {\em Proc.
  NeurIPS}, 2021.

\bibitem{limberg2022yolo}
C.~Limberg, A.~Melnik, A.~Harter, and H.~J. Ritter, ``{YOLO - You only look
  10647 times},'' {\em arXiv preprint arXiv:2201.06159}, 2022.

\bibitem{yolohow1}
P.~Jędrzej~Świeżewski, ``Introduction to yolo algorithm and yolo object
  detection.'' \url{https://appsilon.com/object-detection-yolo-algorithm/}.
\newblock accessed February 27, 2022.

\bibitem{redmon2016look}
J.~Redmon, S.~Divvala, R.~Girshick, and A.~Farhadi, ``You only look once:
  Unified, real-time object detection,'' 2016.

\bibitem{yolohow1_1}
``{YOLOv5}.'' \url{https://github.com/ultralytics/yolov5/}.

\bibitem{AIBlitzXIII-MaskPrediction}
AIcrowd, ``{AI Blitz XIII Face: Mask Prediction - Predict Mask \& Bounding Box
  From Images}.''
  \url{https://www.aicrowd.com/challenges/ai-blitz-xiii/problems/mask-prediction};
  accessed April 4, 2022.

\bibitem{AIBlitzXIII-FaceRecognition}
AIcrowd, ``{AI Blitz XIII Face: Face Recognition - Find The Correct Face In
  Crowd}.''
  \url{https://www.aicrowd.com/challenges/ai-blitz-xiii/problems/face-recognition};
  accessed April 4, 2022.

\bibitem{faceRecogGithub}
GitHub, ``face-recognition.''
  \url{https://github.com/ageitgey/face_recognition}; accessed April 4, 2022.

\bibitem{dlibGithub}
GitHub, ``dlib.'' \url{https://github.com/davisking/dlib}; accessed April 4,
  2022.

\bibitem{faceAligGithub}
A.~Bulat, I.~Toubal, J.-Y. Lee, S.~Bogdanov, J.~J. Lewis, S.~Jadhav, S.~Rezaei,
  P.~Grunt, and A.~Solanki, ``face-alignment.''
  \url{https://github.com/1adrianb/face-alignment}, 2021.

\bibitem{mediapipePaper}
C.~Lugaresi, J.~Tang, H.~Nash, C.~McClanahan, E.~Uboweja, M.~Hays, F.~Zhang,
  C.-L. Chang, M.~G. Yong, J.~Lee, W.-T. Chang, W.~Hua, M.~Georg, and
  M.~Grundmann, ``Mediapipe: A framework for building perception pipelines,''
  tech. rep., Google Research, 2019.

\bibitem{mediapipeGithub}
GitHub, ``Mediapipe.'' \url{https://github.com/google/mediapipe}; accessed
  April 4, 2022.

\bibitem{mediapipeGroupsGithub}
GitHub, ``Mediapipe landmark groups.''
  \url{https://github.com/google/mediapipe/blob/master/mediapipe/python/solutions/face_mesh_connections.py};
  accessed April 4, 2022.

\bibitem{AIBlitzXIII-FaceDeBlurring}
AIcrowd, ``{AI Blitz XIII Face: Face De-Blurring - Convert Blurred image Into
  Clear Image}.''
  \url{https://www.aicrowd.com/challenges/ai-blitz-xiii/problems/face-de-blurring};
  accessed April 4, 2022.

\bibitem{phdthesis}
H.~Yeganeh, {\em Cross Dynamic Range And Cross Resolution Objective Image
  Quality Assessment With Applications}.
\newblock PhD thesis, University of Waterloo, 07 2014.

\bibitem{dong2015image}
C.~Dong, C.~C. Loy, K.~He, and X.~Tang, ``Image super-resolution using deep
  convolutional networks,'' {\em IEEE transactions on pattern analysis and
  machine intelligence}, vol.~38, no.~2, pp.~295--307, 2015.

\bibitem{FacesSurvey}
A.~Melnik, ``{Deep Face Generation and Editing: A Survey}.''
\newblock
  \url{https://github.com/ndrwmlnk/deep-face-generation-and-editing-a-survey};
  accessed April 4, 2022.

\bibitem{mao2021deep}
X.~Mao, Y.~Liu, W.~Shen, Q.~Li, and Y.~Wang, ``Deep residual fourier
  transformation for single image deblurring,'' {\em arXiv preprint
  arXiv:2111.11745}, 2021.

\bibitem{wang2021uformer}
Z.~Wang, X.~Cun, J.~Bao, and J.~Liu, ``Uformer: A general u-shaped transformer
  for image restoration,'' {\em arXiv preprint arXiv:2106.03106}, 2021.

\bibitem{zamir2021restormer}
S.~W. Zamir, A.~Arora, S.~Khan, M.~Hayat, F.~S. Khan, and M.-H. Yang,
  ``Restormer: Efficient transformer for high-resolution image restoration,''
  {\em arXiv preprint arXiv:2111.09881}, 2021.

\bibitem{1292216}
Z.~Wang, E.~Simoncelli, and A.~Bovik, ``Multiscale structural similarity for
  image quality assessment,'' in {\em The Thrity-Seventh Asilomar Conference on
  Signals, Systems Computers, 2003}, vol.~2, pp.~1398--1402 Vol.2, 2003.

\end{thebibliography}
\end{document}